\documentclass[10pt, twocolumn]{article}

\usepackage[letterpaper, margin=1in]{geometry}   

\usepackage{times}
\usepackage{epsfig}
\usepackage{graphicx}
\usepackage{amsmath}
\usepackage{amssymb}
\usepackage{booktabs}
\usepackage[colorlinks=true,breaklinks=true]{hyperref}

\newcommand{\keywords}[1]{\textbf{\emph{Keywords:}} #1}

\title{
DensifyBeforehand: LiDAR-assisted Content-aware Densification\\
for Efficient and Quality 3D Gaussian Splatting
}

\author{
    Phurtivilai Patt \quad
    Leyang Huang \quad
    Yinqiang Zhang \quad
    Yang Lei \\
    {\small The University of Hong Kong} \\
}

\date{}  

\begin{document}

\maketitle

\begin{abstract}
This paper addresses the limitations of existing 3D Gaussian Splatting (3DGS) methods, particularly their reliance on adaptive density control, which can lead to floating artifacts and inefficient resource usage. We propose a novel ``densify beforehand'' approach that enhances the initialization of 3D scenes by combining sparse LiDAR data with monocular depth estimation from corresponding RGB images. Our ROI-aware sampling scheme prioritizes semantically and geometrically important regions, yielding a dense point cloud that improves visual fidelity and computational efficiency. 
This ``densify beforehand'' approach bypasses the adaptive density control that may introduce redundant Gaussians in the original pipeline, allowing the optimization to focus on the other attributes of 3D Gaussian primitives, reducing overlap while enhancing visual quality. 
Our method achieves comparable results to state-of-the-art techniques while significantly lowering resource consumption and training time. We validate our approach through extensive comparisons and ablation studies on four newly collected datasets, showcasing its effectiveness in preserving regions of interest in complex scenes.
\end{abstract}


\section{Introduction}
Neural Radiance Fields~\cite{mildenhall2021nerf, barron2022mip} that can render novel view images of a scene have attracted increasing research interest. Among all relevant works, 3D Gaussian Splatting (3DGS)~\cite{kerbl20233d} uses the sparse points from structure-from-motion (SfM)~\cite{schoenberger2016sfm} when tracking the camera poses of the input multi-view images to initialize a 3D scene and employs a point-based approach that drastically speeds up the rendering efficiency, paving the way for various computer vision and robotic applications. Based on this recent breakthrough, recent works~\cite{keetha2024splatam, matsuki2024gaussian, jiang2024li, hong2024liv, lang2024gaussian, xiao2024liv} accompany the RGB camera with a ranging sensor (e.g., RGB-Depth cameras or LiDAR device) to utilize 3DGS for simultaneous localization and mapping (SLAM) and fast or large-scale scene reconstruction. Compared to using the sparse point cloud produced by SfM, LiDAR sensors can provide a readily measurable point cloud for many downstream tasks, such as construction inspection or progress monitoring~\cite{zhang2024global}.

\begin{figure}
    \centering
    \includegraphics[width=\linewidth]{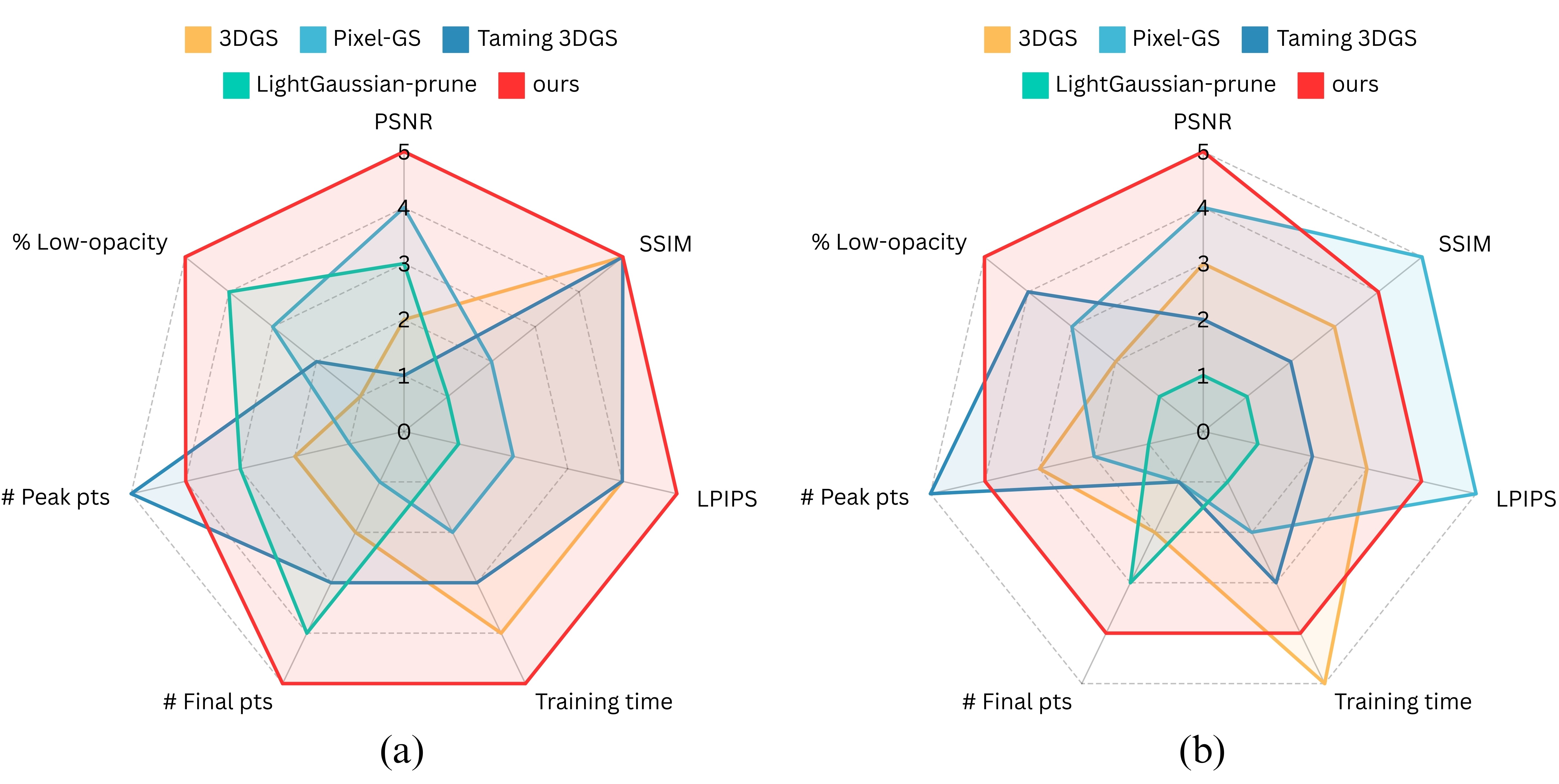}
    \caption{Our performance comparing to other SOTA methods on \textit{wetlab} dataset (a) and \textit{corner} dataset (b). Our method can achieve comparable performance in terms of rendering quality and use fewer Gaussians to represent the same scene.}
    \label{fig:radar-chart}
\end{figure}

With the increasing accessibility of miniaturized LiDAR sensors on personal mobile devices~\cite{baruch2021arkitscenes}, such as the iPhone Pro series, a promising application would be how to use the miniaturized LiDAR sensor to facilitate modeling a scene with 3DGS.
Despite their advantages of being accessible and less sensitive to lighting conditions (compared to RGB-Depth cameras), miniaturized LiDAR sensors usually provide very \textit{sparse} ranging data~\cite{ren2024ags, behari2024blurred} and are with a limited ranging distance, thereby restricting their performance in modeling complex 3D scenes, particularly in regions with fine details or thin objects. 

Previous LiDAR-based works optimize their 3DGS scenes with the adaptive density control (ADC) strategy proposed in~\cite{kerbl20233d} to grow Gaussian primitives by cloning, splitting, and pruning. However, ADC faces several limitations, such as floating artifacts~\cite{zhang2024pixel} and an excessive number of Gaussian primitives~\cite{deng2024efficient}, as discussed in Sec.~\ref{sec:relatedworks}, potentially compromising the overall visual fidelity.

\begin{figure*}[htbp]
    \centering
    \includegraphics[width=1\linewidth]{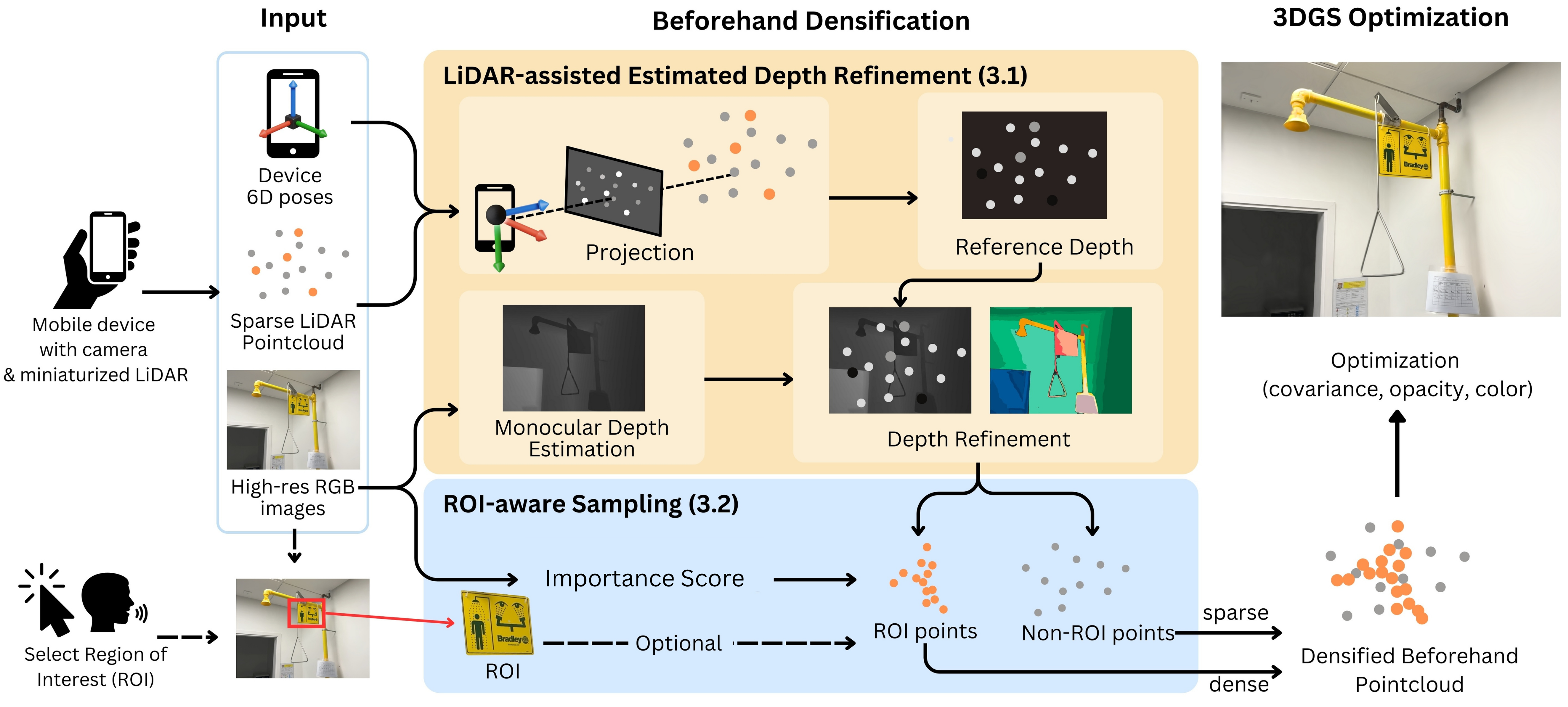}
    \caption{Our ``densify beforehand" approach. Our method takes as input a set of posed RGB frames real-time tracked from a visual-LiDAR device and the resulting sparse point cloud. We adopt a monocular depth estimation (MDE) method to derive dense depth maps from RGB frames and utilize the sparse LiDAR points to rescale the MDE results. To prepare dense initial points for training 3DGS, we adopt an ROI-aware importance sampling strategy. Finally we train the dense input points with 3DGS splitting and pruning to yield an efficient and compact scene.}
    \label{fig:workflow}
\end{figure*}

In this paper, we aim to propose a novel, streamlined approach to 3DGS scene representation. This approach takes as input from a consumer-grade mobile device: 1) the tracked poses of the device, 2) sparse point clouds by a miniaturized LiDAR sensor, and 3) high-resolution RGB images captured by an RGB camera. As shown in Fig.~\ref{fig:workflow}, it then resorts to the monocular depth estimation (MDE) approach (e.g.,~\cite{DBLP:journals/pami/metric3dv2}) for initial point cloud densification.
This ``densify-beforehand" approach uses a (nearly) constant number of Gaussian primitives from the initialization, offering a practical controllability of the optimization process. It also achieves high-quality rendering results without using several million Gaussian primitives, yielding a resource-aware and efficient implementation by leveraging LiDAR input.

We prioritize computational resources by defining the regions of interest (ROI) in the RGB images (such as texture-rich regions or foreground objects and texts/signs) while downsampling other regions. This yields a dense point cloud to represent the ROIs in the given scene. As we will show in our results, our method can model a scene with 3DGS using a much smaller number of Gaussian primitives yet achieving a comparable rendering quality.

Comparison with state-of-the-art methods shows that our ``densify-beforehand'' approach can achieve results comparable to the adaptive density control of the original 3DGS~\cite{kerbl20233d} or an improved version proposed in Pixel-GS~\cite{zhang2024pixel} in terms of visual quality, while drastically reducing resource consumption in terms of the final number of Gaussians and training time. 
We also compare our method with Taming 3DGS~\cite{mallick2024taming} and LightGaussian~\cite{fan2023lightgaussian} to demonstrate our effectiveness in a resource-constrained set-up. Ablation studies are conducted to validate our design choices.
Furthermore, our MDE-based densification pipeline generates dense point clouds comparable to those produced by depth sensors, validating its effectiveness. 
We collected six new datasets featuring scenes with thin geometry objects and rich textual content, to validate the proposed method. We also showcase some applications where our reconstructed scenes can preserve regions of interest specified by users (such as using a verbal description).

In summary, our technical contributions are:

1) A novel offline, content-aware densification approach that bypasses the adaptive density control in the existing 3DGS pipeline.

2) A practical strategy that leverages sparse LiDAR points to refine the monocular depth estimation results for 3DGS scene representation.

3) Six datasets demonstrating a diverse range of appearances captured using a camera and a miniaturized LiDAR sensor.

\section{Related works}\label{sec:relatedworks}

\textbf{LiDAR-assisted 3DGS. }
The accessibility of LiDAR sensors and recent progress in multi-sensor fusion~\cite{zheng2024fast, hong2024liv} lead to increasing adoption of visual-LiDAR systems in reconstructing 3D scenes~\cite{hong2024liv, lang2024gaussian, xiao2024liv, cui2024letsgo}.
In contrast to these previous works that utilize high-end LiDAR sensors, our method focuses on a miniaturized LiDAR equipped on a commodity-grade mobile device. Since a miniaturized LiDAR sensor produces a very sparse point cloud, we propose to leverage the recent advancement in monocular depth estimation (MDE), such as~\cite{DBLP:journals/pami/metric3dv2, DBLP:conf/nips/depthanythingv2, DBLP:journals/corr/depthpro} to name just a few, to densify the sparse LiDAR point cloud. We show that this approach can bypass the error-prone yet expensive adaptive density control, allowing our method to achieve SOTA performance in terms of both visual quality and efficiency.

\textbf{Adaptive Density Control. }
Original 3DGS employs adaptive density control to grow the 3D scene initialized from SfM. However, this adaptive density control strategy suffers from several limitations that may yield blurry rendered images of the scene. Large Gaussian primitives can be produced due to the sparse, unevenly distributed initial points produced by SfM. Pixel-GS~\cite{zhang2024pixel} addresses this limitation by taking into account the number of pixels a Gaussian primitive covers in each view in the optimization. This weighted average formulation encourages the point growth in regions covered by large Gaussian primitives, thus avoiding the blurry results due to the presence of large Gaussian primitives.
Unlike these methods that rely on the \textit{sparse point cloud} from SfM or the \textit{adaptive density control} (ADC), we leverage a mobile visual-LiDAR device (i.e., an iPhone Pro 16) and propose a novel approach that densifies the scene before optimization. 
In particular, we replace the cloning operation, one of the three major components of the ADC that increases Gaussians in the scene, with the proposed DensifyBeforehand point cloud. 
We validate our design choice by comparison with Pixel-GS~\cite{zhang2024pixel} and the original 3DGS~\cite{kerbl20233d} that use the same LiDAR point cloud for initialization.

\textbf{Efficiency. }
Foreground objects are contextually important for the scene and may require a large number of 3D Gaussian primitives to model in the original pipeline.
However, representing a scene with an excessive number of Gaussian primitives requires substantial GPU memory for rendering and impairs real-time rendering performance.
Tamining 3DGS~\cite{mallick2024taming} proposes a score-based densification approach that modifies the original adaptive density control to be resource-aware. 

Our method adopts a simple importance sampling approach to distribute more sampled points in the region of interest (where color variance is high by default). This approach allows a 3DGS scene to be trained with a controllable resource in minutes while maintaining the visual quality of foreground objects or even the texts/signs on them during runtime rendering.
We show that our method can be comparable with Tamining 3DGS~\cite{mallick2024taming} in terms of the visual quality of the rendering outputs while achieving higher training and rendering efficiency.

Observing the 3DGS scene representation contains many Gaussian primitives with very low opacity, recent works also proposed different approaches to cull these low-opacity Gaussians~\cite{fan2023lightgaussian, papantonakis2024reducing}. As shown in the experiments, our 3DGS scenes can be pruned straightforwardly by thresholding the opacity of Gaussians, yet achieve comparable results as the state-of-the-art.

\section{Methodology}\label{sec:method}
3D Gaussian Splatting (3DGS) represents a scene as a collection of 3D Gaussian primitives, centered at $\mu$ with a scaling $S$ and rotation $R$~\cite{kerbl20233d}. The color rendered on the image plane is blended from multiple Gaussian primitives depending on the opacity. The optimization pipeline consists of three major operations: cloning, splitting, and pruning. We propose a densify-beforehand approach instead of performing cloning operations during the optimization.

The overview of our method is shown in Fig.~\ref{fig:workflow}.
Given a set of posed RGB images and a LiDAR point cloud, our method first fuses the sparse LiDAR data and the monocular depth estimation (MDE) results from~\cite{DBLP:journals/pami/metric3dv2} to densify the scene. 
We sample an ROI-aware subset from this densified point cloud, then initialize and train 3DGS with these ROI-aware samples, bypassing the adaptive density control in the original pipeline. 

\subsection{LiDAR-assisted Estimated Depth Refinement}
Given the huge progress in monocular depth estimation~\cite{DBLP:journals/pami/metric3dv2, DBLP:journals/corr/depthpro, DBLP:journals/corr/zoedepth, DBLP:conf/nips/depthanythingv2}, accurately estimating the metric depth of a given RGB image is still challenging. Therefore, to provide a dense point cloud to initialize the 3DGS, we leverage sparse point clouds from a LiDAR sensor to refine the estimated depth from MDE. 

The input to our method contains a set of RGB images $\{I_k\}$ with corresponding poses $\{T_k\}$ derived from LiDAR-based camera tracking, and a point cloud $X = \{x_i\}$. 
We adopt Metric3D-V2-vits-giant~\cite{DBLP:journals/pami/metric3dv2} to provide a per-pixel depth map $D$ for each RGB image $I$. Then, we propose a global-local approach to refine the estimated depth maps $\{D_k\}$ as described.

First, we project the LiDAR points $X$ to each image plane defined by a $T_k$ with the use of hidden point removal~\cite{katz2007direct} to avoid projecting occluded points onto the image plane, denoted $A_k$. Value at each pixel of $A_k$ is either $0$ or the depth from the projected point and $\mathbf{1}_k$ is an indicator function of the non-zero elements of $A_k$. We term the non-zero pixels in $A_k$ as anchors.

\textbf{Global scaling operation.}  We then calculate a global rescaling factor for each $D_k$ based on the anchors in $A_k$:
\[
    g_k = \textsc{Median}\Big( \frac{\mathbf{1}_k * A_k}{\mathbf{1}_k * D_k} \Big),
\]
where $\textsc{Median}$ is the median operator and $*$ denotes the element-wise multiplication.
Thus, the scaled depth map is calculated as 
\begin{equation}\label{eq:global_scaling}
D^{t}_k = g^t_k D^{t-1}_k.
\end{equation}
We perform this global scaling iteratively to ensure the convergence of the scaled depth map, which is usually required no more than 5 iterations, and denote the resulting rescaled depth maps as $\tilde{D}_k$.

\textbf{Local cluster-based scaling operation.}  Upon the convergence of the iterative global scaling operations, multiple scans stack in the scene and point cloud consolitation is need. In order to address this issue, we propose a local cluster-based scaling operation to consolidate the depth maps from MDE. 

We use the anchor pixels \(\mathbf{p}_i\) in $A_k$ to perform clustering on the corresponding monocular depth image $\tilde{D}_k$. The clustering process assigns each pixel $\mathbf{q}_j$ in $\tilde{D}$ to the nearest anchor pixel from \(\mathbf{p}_i\) based on their Euclidean distance
\[
d(\mathbf{p}_i, \mathbf{q}_j) = \big\|(\mathbf{p}_i, A(\mathbf{p}_i)) - (\mathbf{q}_j, \tilde{D}(\mathbf{q}_j))\big\|_{2}.
\]
Since LiDAR points are distributed in the near scene and cannot capture far-away regions, we intentionally filter out any pixels $\mathbf{q}_j$ whose distance to $\mathbf{p}_i$ is larger than a threshold $\tau$ to avoid wrongly rescaling far-away points in $\tilde{D}$. 
Thus, a cluster at $\mathbf{p}_i$ is denoted $\mathcal{C}_i = \{\mathbf{q}_j | d(\mathbf{p}_i, \mathbf{q}_j) < \tau \}$. Within each cluster $\mathcal{C}_i$, we first compute a local scaling factor $\gamma_i$ as 
$\gamma_i = A(\mathbf{p}_i)/\tilde{D}(\mathbf{p}_i)$
and then apply $\gamma_i$ to all pixels $\mathbf{q}_j$ in $\mathcal{C}_i$: 
\begin{equation}\label{eq:local_scaling}
   \hat{D}(\mathcal{C}_i) = \gamma_i \tilde{D}(\mathbf{q}_j \in \mathcal{C}_i).
\end{equation}

\subsection{ROI-aware Sampling for 3DGS}

Dense depth maps of the input RGB frames generated in the previous stage not only densify the sparse LiDAR points $X$ in the foreground scene but also the far-away background scene that cannot be captured by the miniaturized LiDAR. 
However, converting all dense depth maps\footnote{2.7 million pixels per frame for a $1920\times1440$ RGB image} into a point cloud for training would entail a huge computational overhead. Therefore, we subsample a subset of pixels from each image and convert them into the initial point cloud (several hundred thousand points) for 3DGS training.

We employ an importance sampling approach to better reconstruct the region of interest of the scene. Specifically, we aim to define the importance scores to holistically reflect the importance of 3D regions instead of their 2D projections in each RGB frame. Therefore, we accumulate the spatial importance score $s(X)$ from pixels. Given the definition of a per-pixel importance score, we back-project the per-pixel importance scores $s(\mathbf{q}_j)$ to the LiDAR point cloud $X$ for establishing the spatial importance in the 3D scene. 
First, we compute each anchor's importance score $s(\mathbf{p}_i)$ by averaging the importance score of the associated pixels in its corresponding cluster $\mathcal{C}_i$. 
Recalling that anchors are 2D projections of the LiDAR points, we average the importance scores from different anchors of the same LiDAR point to be the importance score $s(x_i \in X)$ of these LiDAR points. Thus, the spatial importance is established.
By default, we use the color variance with a kernel size of 9 to compute the per-pixel importance.

When sampling a dense set of points from 3D, we project the importance scores from the LiDAR point to its corresponding anchor in a certain RGB frame and perform the importance sampling based on this importance score. This treatment ensures a consistent important score within a small spatial neighborhood, avoiding a view-dependent importance scoring scheme and respect both the texture and geometry of the 3D scene. This approach naturally concentrates computing resources (i.e., the samples) on the foreground scene as far-away regions have fewer LiDAR points. 

Given the spatial importance score $s(x)$ in 3D, we threshold it with a pre-defined ratio to the median of the importance score, producing an importance mask. We then project it to all RGB frames $I_k$ with their poses $T_k$. This projected mask is denoted $M_k$ for $k$-th RGB frame.  We generate a point cloud \(P \in \mathbb{R}^{N \times 3}\) from $\hat{D}_k$ using their camera poses $T_k$. The density of the point cloud from ROI or non-ROI regions are controlled by the mask $M_k$. To achieve this, we apply a density parameter \(\rho\) that controls the number of points generated in ROI and non-ROI regions. Specifically, for ROI regions, we use a high sampling density \(\rho_{\text{ROI}}\) to ensure dense point coverage; on the other hand, we use a low sampling density \(\rho_{\text{non-ROI}}\) to maintain computational efficiency. This sampling density ratio ($\rho_{ROI}/\rho_{non-ROI}$) is set to 30 and the masking threshold for defining the ROI is set to the median of the spatial importance scores. We found that putting more points in the ROI regions, or the texture-rich regions, is beneficial to obtain higher reconstruction quality metrics (e.g., PSNR).

The resulting point cloud \(\mathbf{P}\) is used as input for training the 3D Gaussian model. By preprocessing the scene to prioritize ROIs, we ensure that the Gaussian model focuses its resources on ROI areas, leading to more efficient training and accurate reconstruction in these areas. 
This approach avoids the cloning operation in the original ADC pipeline to excessively introduce nearby Gaussians that may be redundant at this end.

\section{Experimental results}\label{sec:results}
In this section, we first introduce our self-collected LiDAR dataset in Sec.~\ref{sec:dataset}. Sec.~\ref{sec:comparison} compares our DensifyBeforehand approach with state-of-the-art methods, demonstrating its comparable performance in visual quality and superiority in training efficiency, storage, and the peak number of Gaussian primitives during optimization. These results demonstrate the practical usefulness of our method. Ablation studies are conducted in Sec.~\ref{sec:ablation} to analyze the contribution of each proposed component.

\subsection{Dataset preparation}\label{sec:dataset}

Six real-world datasets (see Fig.~\ref{fig:gallery}) were collected to facilitate a comprehensive evaluation of our work. The datasets include \textit{Wetlab} (W1), \textit{Reception} (R2), \textit{Corner} (C3), \textit{Pantry} (P4), \textit{Machines} (M5), and \textit{Staircase} (S6). 
An iPad Pro with a miniaturized LiDAR sensor was used for data collection. ARKit~\footnote{https://developer.apple.com/augmented-reality/arkit/} was used to acquire the pose of the sensors and the data. RGB frames are at a resolution of $1920 \times 1440$. See Tab.~\ref{tab:statistics} for more statistics about the datasets.

\begin{table}[]
    \centering
    \resizebox{0.75\linewidth}{!}{    
    \begin{tabular}{c|c c c c c c}
         & W1 & R2 & C3 & P4 & M5 & S6 \\
    \hline
    \#Frames & 191& 321& 564& 156& 285& 249\\
    \#Points &  148,414& 111,710& 177,747& 184,685& 123,247& 218,659\\
    \hline
    \end{tabular}
    }
    \caption{Statistics of the collected datasets. W1: Wet Lab; R2: Reception; C3: Corner; P4: Pantry; M5: Machines; S6: Staircase}
    \label{tab:statistics}
\end{table}

\begin{table*}[]
    \centering
    \resizebox{\linewidth}{!}{
    \begin{tabular}{c|r r r r r r r}
         & PSNR $\uparrow$ & SSIM $\uparrow$ & LPIPS $\downarrow$ & Time (s) $\downarrow$& \#G (k) $\downarrow$& Peak \#G (k)$\downarrow$& $P_{Opaq\leq0.1}$ $\downarrow$ \\
    \midrule
    \multicolumn{8}{c}{\textbf{Wetlab}} \\
    \hline
    3DGS~\cite{kerbl20233d} & 28.8505& 0.9387& 0.1459& 407.61& 918.23& 918.23& 60.1\\
    Pixel-GS~\cite{zhang2024pixel} & 29.6315& 0.9369& 0.1500& 2019.86& 1439.35& 1439.35& 54.17\\
    Taming-GS~\cite{mallick2024taming} & 28.7854& 0.9387& 0.1456& 496.04& 406.86& \textbf{406.86}& 57.91\\
    LightGS~\cite{fan2023lightgaussian}& 28.9727& 0.9362& 0.1580& 2736.09& 285.42& 739.44& 37.83\\  
    Ours & \textbf{29.7187}& \textbf{0.9389}& \textbf{0.1454}& \textbf{402.87}& \textbf{252.90}& 630.71& \textbf{5.52}\\
    \midrule
    \multicolumn{8}{c}{\textbf{Reception}} \\
    \hline
    3DGS~\cite{kerbl20233d} & 25.9346& 0.9017& 0.1643& 465.68& 975.12& 975.12& 55.55\\
    Pixel-GS~\cite{zhang2024pixel} & 25.9772& 0.9016& 0.1657& 1621.12& 1678.7& 1678.7& 41.94\\
    Taming-GS~\cite{mallick2024taming} & \textbf{26.0844}& \textbf{0.9029}& \textbf{0.1641}& 524.18& 474.12& \textbf{474.12}& 51.64\\
    LightGS~\cite{fan2023lightgaussian} & 25.8042& 0.8964& 0.1877& 2820.87& 474.14& 745.09& 41.91\\
    Ours & 25.8380& 0.8997& 0.1656& \textbf{440.01}& \textbf{333.59}& 649.81& \textbf{4.28}\\
    \midrule
    \multicolumn{8}{c}{\textbf{Corner}} \\
    \hline
    3DGS~\cite{kerbl20233d} & 28.2961& 0.9096& 0.1553& \textbf{559.92}& 1847.52& 1847.52& 62.51\\
    Pixel-GS~\cite{zhang2024pixel} & 28.4304& \textbf{0.9134}& \textbf{0.146}& 2076.66& 3629.32& 3636.69& 60.93\\
    Taming-GS~\cite{mallick2024taming} & 28.145& 0.9043& 0.1648& 720.00& \textbf{521.24}& \textbf{985.17}& 57.56\\
    LightGS~\cite{fan2023lightgaussian}* & 27.4819& 0.8949& 0.1850 & 3316.4& 1095.78& 3025.00 & 22.82 \\

    Ours & \textbf{28.4400}& 0.9110& 0.1500& 577.74& 666.20& 1498.87& \textbf{4.54}\\

    \midrule
    \multicolumn{8}{c}{\textbf{Machines}} \\
    \hline
    3DGS~\cite{kerbl20233d} & 24.422& 0.8111& 0.2696& 480.81& 1349.9& 1349.9& 56.52\\
    Pixel-GS~\cite{zhang2024pixel} & 24.5386& 0.8143& \textbf{0.2594}& 1927.67& 2970.21& 2970.21& 45.81\\
    Taming-GS~\cite{mallick2024taming} & \textbf{24.6384}& \textbf{0.8154}& 0.2633& 661.42& 999.96& \textbf{999.96}& 59.65\\
    LightGS~\cite{fan2023lightgaussian} & 23.8544& 0.7965& 0.3044& 2758.96& 1000.00& 1104.82& 57.27\\
    Ours & 24.4118& 0.8103& 0.2667& \textbf{454.88}& \textbf{532.57}& 1099.45& \textbf{3.95}\\
    \midrule
    \multicolumn{8}{c}{\textbf{Pantry}} \\
    \hline
    3DGS~\cite{kerbl20233d} & 27.2479& 0.9063& 0.1734& 382.11& 863.09& 863.09& 56.29\\
    Pixel-GS~\cite{zhang2024pixel} & \textbf{27.5212}& \textbf{0.9078}& \textbf{0.1710}& 1524.78& 1554.57& 1554.57& 41.38\\
    Taming-GS~\cite{mallick2024taming} & 27.3060& 0.9060& 0.1739& 448.44& 446.25& \textbf{446.25}& 54.05\\
    LightGS~\cite{fan2023lightgaussian} & 27.2005& 0.9048& 0.1864& 2638.26& 446.26& 723.69& 34.08\\
    Ours & 26.9792& 0.9023& 0.1777&\textbf{ 379.42}& \textbf{330.57}& 699.79 & \textbf{6.14}\\
    \midrule
    \multicolumn{8}{c}{\textbf{Staircase}} \\
    \hline
    3DGS~\cite{kerbl20233d} & 27.8365& 0.8904& 0.1912& 524.63& 1349.62& 1349.62& 57.75\\
    Pixel-GS~\cite{zhang2024pixel} & \textbf{27.8628}& 0.891& \textbf{0.1875}& 1869.42& 2404.83& 2404.83& 54.41\\
    Taming-GS~\cite{mallick2024taming} & 28.0503& \textbf{0.8927}& 0.1915& 542.22& 581.38& \textbf{581.38}& 52.02\\
    LightGS~\cite{fan2023lightgaussian} & 27.2810& 0.8809& 0.2157& 2572.47& 581.40& 1125.02& 33.22\\
    Ours & 27.7073& 0.8875& 0.1940& \textbf{433.73}& \textbf{385.42}& 877.40& \textbf{5.30}\\
    \hline
    \end{tabular}
    }
    \caption{
    Quantitative comparison with SOTA on six self-collected datasets. All comparing methods are initialized with the same LiDAR point cloud for fair comparison. Our method densifies the LiDAR point cloud with the use of a monocular depth estimation method. Besides the metrics for rendering quality, we also report Time (s) for training, the number of final/peak Gaussians in thousands (\#G / Peak\#G), and the percentage of Gaussians with opacity lower than 0.1. 
    *LightGS cannot fit in NVIDIA RTX4090 GPU and thus is trained on L20 (with 48GB GPU memory) which is slower.}
    \label{tab:sota_comparison}
\end{table*}

\subsection{Comparison with SOTA}\label{sec:comparison}

\textbf{Comparing methods. }
To validate our approach, we compared our method against two groups of state-of-the-art (SOTA) methods. The first group is the original 3DGS~\cite{kerbl20233d} and Pixel-GS~\cite{zhang2024pixel} that uses a huge number of Gaussians to represent the scene. The other group contains LightGaussian~\cite{fan2023lightgaussian} and Taming 3DGS~\cite{mallick2024taming} that improve 3DGS's computational overhead.

\textbf{Implementation details.}
All experiments were conducted using PyTorch~\cite{DBLP:conf/nips/PaszkeGMLBCKLGA19} on a desktop computer equipped with an NVIDIA RTX4090 GPU (24GB) and an Intel Core i9-12900K CPU running on Linux Ubuntu 22.04. 

We implemented our work based on the original 3DGS framework~\cite{kerbl20233d}. All comparing methods~\cite{mallick2024taming, zhang2024pixel, fan2023lightgaussian} were implemented with their open-sourced codes and default hyperparameters.
We downsampled the input images to a resolution of 1600 $\times$ 1200 to allow all methods, including the more computationally intensive ones, to be trained with an NVIDIA RTX4090 GPU.
All timings start at loading the dataset and conclude once the final checkpoint is saved. 

\textbf{Metrics.}
We report established metrics from previous studies, i.e., PSNR, SSIM, and LPIPS to evaluate the quality of the rendering results. Additionally, we also compare the training time in seconds (s), as well as the final and peak numbers of Gaussian primitives during training (denoted \#G and Peak \#G, respectively). These metrics serve as indicators of computational efficiency and overhead of the comparing methods.

\begin{figure*}[!t]
    \centering
    \includegraphics[width=1\linewidth]{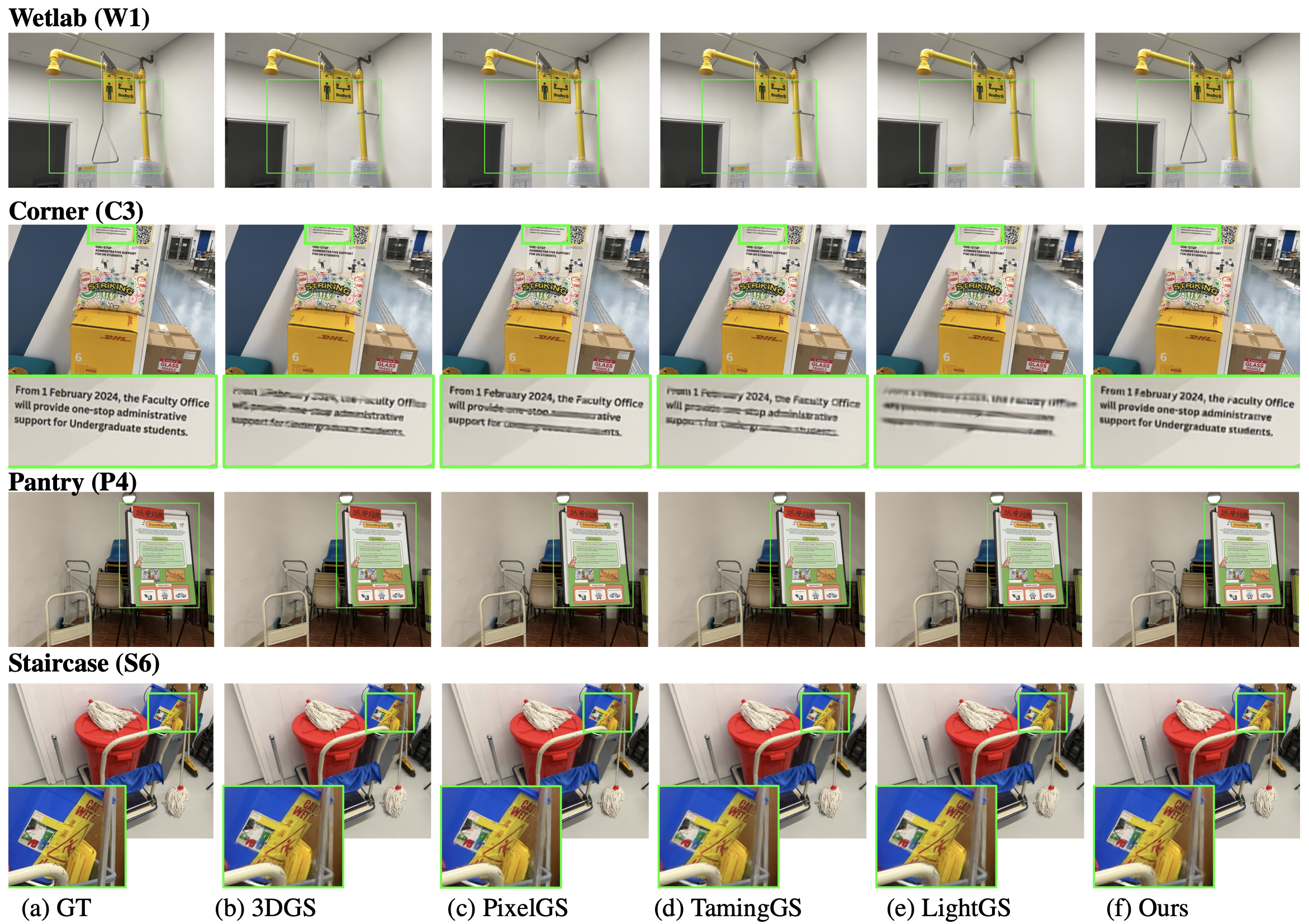}
    \caption{Qualitative comparisons of our results and those produced by the SOTA methods. Our method can capture the puller in the Wetlab scene (in the first row) and reconstruct both the text (enclosed) and the far-away scene of the Corner scene (in the second row). We achieve comparable visual quality with the comparing methods in the Pantry (third) and Staircase (fourth) where PixelGS excels in capturing the texts.}
    \label{fig:gallery}
\end{figure*}

\textbf{Results. }
Tab.~\ref{tab:sota_comparison} quantitatively compares the novel-view rendering quality of our methods to those of the SOTA methods. We also facilitate the readers to appreciate our method's advantages and limitations by a qualitative comparison with the SOTA methods in Fig.~\ref{fig:gallery}.

First, we compare our results with those produced by 3DGS~\cite{kerbl20233d} on the datasets we collected. We observe that our results are comparable to, if not better than, the 3DGS results. Fig.~\ref{fig:gallery} shows that with our densify-beforehand approach, our 3DGS scene can eventually capture the puller in the scene which is thin and with a complicated shape. On the other hand, the original 3DGS pipeline can capture only the upper part of the puller which is straight; the lower part with a triangular shape is challenging for the original pipeline as cloning and splitting the initial points (sparse LiDAR points) cannot effectively move existing Gaussians to this complicated region.
Comparing our method with 3DGS in terms of the computational overhead also demonstrates our advantages. Our method requires less training time and produces consistently fewer Gaussians at the end across different scenes than the 3DGS original pipeline with cloning. Despite using less resources, our method can represent the scene with a slightly higher reconstruction quality on Wetlab (W1) and Corner (C3) datasets and achieve comparable performance on Staircase (S6) and Pantry(P4) datasets.
Similar to 3DGS, PixelGS~\cite{zhang2024pixel} also entails a huge number of Gaussians (around 2.5 million) to represent a scene with comparable visual quality (see Fig.~\ref{fig:gallery}(b) and Tab.~\ref{tab:sota_comparison} Staircase), entailing a prolonged training time.

Our approach can be seen as an alternative to the original 3DGS pipeline and Taming3DGS where a moderate size of Gaussian primitives are initialized in a scene for the optimizer to grow. With the monocular depth estimation results (rescaled by the LiDAR input as reference), we approach the 3DGS initialization more aggressively by putting a large number of points at the beginning and letting the optimizer prune them. We show that this proposed approach can achieve comparable results with Taming3DGS, with better performance on Wetlab (W1) and Corner (C3) and slightly failing behind Taming3DGS on Reception (R2) and Staircase (S6). In terms of computational overhead, our method consistently outperforms Taming3DGS in the training time and our final results can be as compact as 200,000 Gaussians.

LightGS~\cite{fan2023lightgaussian} is designed to derive a more compact 3DGS representation, which requires a trained 3DGS as input. We followed their source codes for implementation and observed that LightGS sometimes incurs a high usage of GPU memory and failed to train the scene Corner (C3) on the RTX4090 GPU. We have to report results regarding this particular scene on a GPU server. In contrast, with our DensifyBeforehand approach, our method can strictly control the number of Gaussians and thus the GPU memory usage. 

From Tab.~\ref{tab:sota_comparison}, we can see that our method constantly outperforms LightGS with higher reconstruction metrics with a lower number of less Gaussian primitives.

The difference between our method and Taming3DGS or LightGS lies in how the scoring function is implemented. While they incorporate a scoring function in the optimization loop to compress the final 3DGS scenes, we opt to spatially score the scene at the initialization. Therefore, we can furnish as many initial candidates as possible for the splitting and pruning operations to optimize the initial Gaussians. Since the dense initialization, we observed that our scenes usually contain a large number of very low-opacity Gaussians and can be straightforwardly removed by setting a threshold. Thus, our approach avoids a sophisticated algorithmic design and parameter search for reaching to a certain number of final Gaussians as LightGS does. We can apply a straightforward pruning at the 25,000-th iteration of our optimization to remove any Gaussians with an opacity lower than 0.1. This simplicity demonstrates the practical advantage of our proposed method.

\begin{figure}
    \centering
    \includegraphics[width=\linewidth, trim = 0 100 0 0, clip]{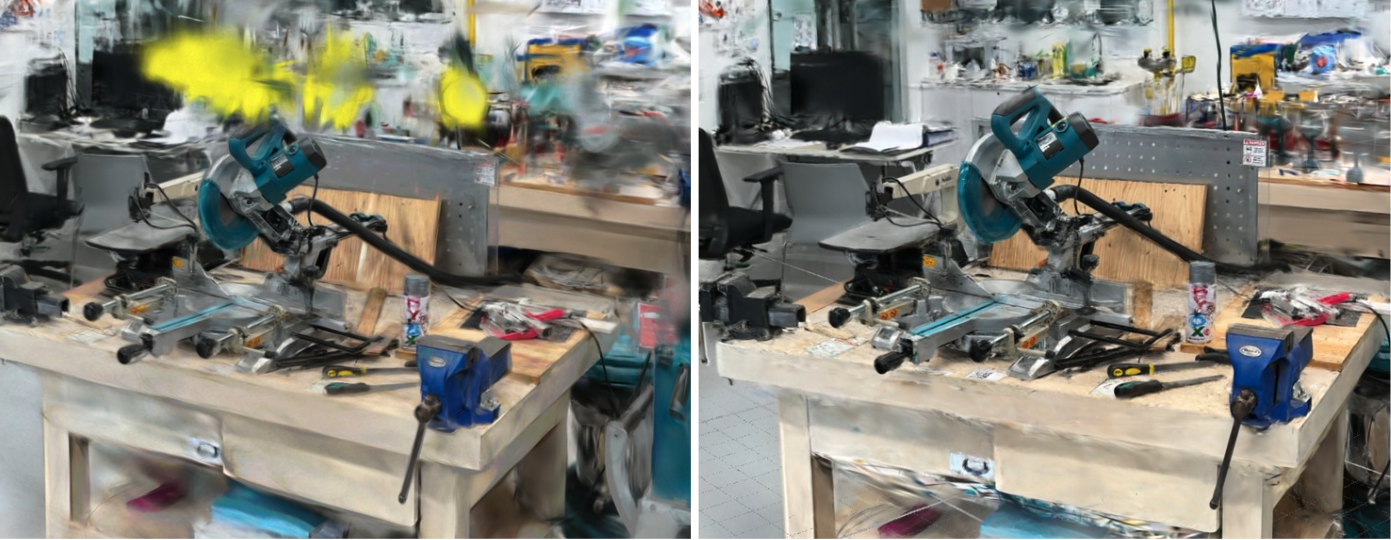}
    \caption{Novel-view renderings to showcase floating artifacts. Left: LightGS, Right: Our result.}
    \label{fig:novelview}
    \vspace{-4mm}
\end{figure}

The proposed method can consistently avoid excessive floating artifacts. This is exemplified in Figure~\ref{fig:novelview}, where a non-test view is rendered by LightGS~\cite{fan2023lightgaussian} (left) and our method (right) are compared. We highlight the floaters (in yellow) above the machines in LightGS and we do not observe floaters around the tabletop.

\subsection{Ablation and discussion}\label{sec:ablation}

We first perform an ablation study on the ROI-aware importance sampling. We report the results on Corner (C2) dataset in Tab.~\ref{tab:ablation-preprocess}. Our method and the ablated version using uniform sampling are given around 1.5 million Gaussians as the budget. The proposed ROI-aware importance sampling generates a dense point cloud from the depth maps, which can lead to higher performance in all three metrics. This validates the use of ROI-aware importance sampling.
We also ablate our method without the late pruning at 25,000 iterations and show that this straightforward pruning operation does not bring any adversary effect to our method.

\begin{table}[]
    \centering
    \resizebox{\linewidth}{!}{
     \begin{tabular}{l|c c c | c }
         & PSNR $\uparrow$ & SSIM $\uparrow$ & LPIPS $\downarrow$ & \#G (k) $\downarrow$\\
    \hline
    Ours (ROI-aware) & 28.5410 & 0.9109& 0.1490 & 666 \\
    Uniform Sampling & 28.3167& 0.9095& 0.1490 & 698 \\
    No Pruning at 25k & 28.5463 & 0.9113 & 0.1467 & 1,344 \\
    \hline
    \end{tabular}
    }
    \caption{Ablation study on Beforehand Densification on the \textit{Corner} (C2) dataset.
    }
    \label{tab:ablation-preprocess}
\end{table}

We also examine how the estimated depth influences our results. Therefore, we compare our results produced by the ground-truth depth maps and depth maps from different sizes of the Metric3D-V2 (small, large, or giant) on the ICL living room dataset where the ground-truth is available. Tab.~\ref{tab:ablation-densification} demonstrates that the rendering quality in term of the three metrics produced by each method varies little, showing the robustness of the proposed method in leveraging MDE methods' results for building 3DGS scenes. The runtime for Metric3Dv2-small and giant to estimate the depth maps from 200 RGB images at resolution $1920\times1440$ is around 1.5 mins and 5 mins, respectively.

\begin{table}[]
    \centering
    \resizebox{\linewidth}{!}{
        \begin{tabular}{l|c c c c c c}
            & PSNR $\uparrow$ & SSIM $\uparrow$ & LPIPS $\downarrow$ & Time (s) $\downarrow$  & \#G (k) $\downarrow$ &$P_{Opaq\leq0.1}$ $\downarrow$\\
            \hline
            100\% & 31.726 & 0.9204 & 0.2479 & 258.45 & 445 & 5.13\\
            20\% + M3D-S & 31.8126 & 0.9212 & 0.2471 & 263.77 & 375 & 4.88\\
            20\% + M3D-L & 31.8459 & 0.9213 & 0.2483 & 262.31 & 370 & 4.83\\
            20\% + M3D-G & 31.8018 & 0.9212 & 0.2478 & 261.02 & 372 & 4.73\\
            \hline
        \end{tabular}
    }
    \caption{Validation of our method using different sizes of Metric3D-V2 models and ground-truth depth on the synthetic \textit{ICL living room} dataset.}
    \label{tab:ablation-densification}
\end{table}

\begin{figure}
    \centering
    \includegraphics[width=1\linewidth]{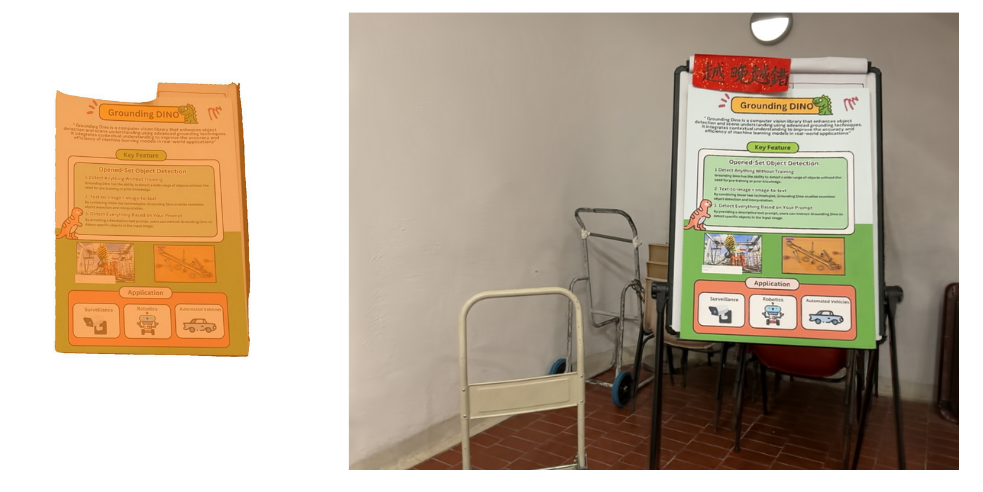}
    \caption{User-specified semantic mask and improved readability of the poster in the Staircase scene.}
    \label{fig:sam}
\end{figure}

We also showcase an application in Fig.~\ref{fig:sam} where the SAM can be applied to specify an ROI which further improves the readability of the poster shown in the Staircase scene.

\subsection{Limitations. }
While our method achieves comparable results with SOTA methods with a lower budget of Gaussian primitives, we also point out some limitations of the proposed method. First, our method relies on the monocular depth estimation algorithms which may provide inaccurate results, especially for texture-rich objects (e.g., posters) in our datasets. While our global-local scaling operations can refine the depth maps in a degree, this will lead to floaters. Second, we currently utilize the color variance as the indicator of default ROI. In the future we will explore a better way to support users to define the ROI and automatically distribute the samples on-the-fly during 

\section{Conclusion}\label{sec:conclusion}

In this paper, we propose a Densify-Beforehand approach to prepare dense initial point cloud for 3DGS representation by leveraging the monocular depth estimation and the sparse LiDAR point cloud provided by a miniaturized LiDAR. 
This is to circumvent the limitations of existing adaptive density control used in the original 3D Gaussian Splatting, in particular, to avoid excessively introducing points by the cloning operation. 
We compare the proposed method to state-of-the-art methods, such as 3DGS, PixelGS, Taming3DGS, and LightGaussian, and demonstrate the effectiveness of the proposed method in reconstructing the scene as well as advantage in computational efficiency. We ablated the proposed method to validate our design choices. We showcase our results in six diverse scenarios.


{\small
\bibliographystyle{cvm}
\bibliography{cvmbib}
}

\end{document}